\documentclass{article}
\usepackage{spconf,amsmath,epsfig}
\usepackage{color, colortbl}
\usepackage{adjustbox}
\usepackage{booktabs}
\usepackage{multirow}
\usepackage{amssymb}
\usepackage{subfigure}
\usepackage{parskip}
\usepackage{bbm}
\RequirePackage{makecell}

\let\OLDthebibliography\thebibliography
\renewcommand\thebibliography[1]{
  \OLDthebibliography{#1}
  \setlength{\parskip}{0pt}
  \setlength{\itemsep}{0pt plus 0.3ex}
}

\begin{document}\sloppy

\def\x{{\mathbf x}}
\def\L{{\cal L}}

\title{Cross-Modal Adapter: Parameter-Efficient Transfer Learning Approach for Vision-Language Models}
\name{Juncheng Yang, Zuchao Li, Shuai Xie, Weiping Zhu, Wei Yu, Shijun Li}
\address{}

\maketitle

\begin{abstract}
Adapter-based parameter-efficient transfer learning has achieved exciting results in vision-language models. Traditional adapter methods often require training or fine-tuning, facing challenges such as insufficient samples or resource limitations. While some methods overcome the need for training by leveraging image modality cache and retrieval, they overlook the text modality's importance and cross-modal cues for the efficient adaptation of parameters in visual-language models. This work introduces a cross-modal parameter-efficient approach named XMAdapter. XMAdapter establishes cache models for both text and image modalities. It then leverages retrieval through visual-language bimodal information to gather clues for inference. By dynamically adjusting the affinity ratio, it achieves cross-modal fusion, decoupling different modal similarities to assess their respective contributions. Additionally, it explores hard samples based on differences in cross-modal affinity and enhances model performance through adaptive adjustment of sample learning intensity. Extensive experimental results on benchmark datasets demonstrate that XMAdapter outperforms previous adapter-based methods significantly regarding accuracy, generalization, and efficiency.

\end{abstract}
\section{Introduction}
\label{sec:intro}
Vision-Language Models (VLM), such as CLIP~\cite{radford2021learning}, ALIGN~\cite{jia2021scaling}, and BLIP~\cite{li2022blip}, have demonstrated excellent performance across multiple tasks, including recognition, generation, and classification.
Traditional pre-training models achieve adaptation to downstream tasks by introducing various auxiliary losses and fine-tuning all parameters of the model. This approach generates a separate set of fine-tuning parameters for each task, proving effective on smaller-scale models. However, these methods suffer from two drawbacks: (i) As the model size increases, fine-tuning requires more significant time and space resources. (ii) For VLM pre-trained on millions of image-text pairs, this approach may lead to overfitting in scenarios with a limited number of available samples. To address these issues, efficient transfer learning (ETL) proposes fine-tuning a small subset of parameters. This allows for the transfer of task-relevant knowledge from the VLM to downstream tasks while mitigating the challenges associated with increased model scale and resource constraints.

Common ETL methods include LoRA~\cite{hu2021lora}, Prompt~\cite{lester2021power}, and Adapter~\cite{houlsby2019parameter}.
LoRA~\cite{hu2021lora} adopts a low-rank matrix decomposition approach by introducing an auxiliary matrix alongside the original matrix. The model updates are achieved by modifying this auxiliary matrix. This ETL method tends to yield moderate performance on smaller-scale models.
CoOp~\cite{zhou2022learning} first proposes turning the fixed prompt ``this is a photo of a [CLS]'' in CLIP~\cite{radford2021learning} into a learnable prompt, enhancing the model's generalization capabilities. However, this type of prompt is sensitive to parameters and can be challenging to train.

\begin{figure}[t] 
\centering
    \includegraphics[width=0.74\linewidth]{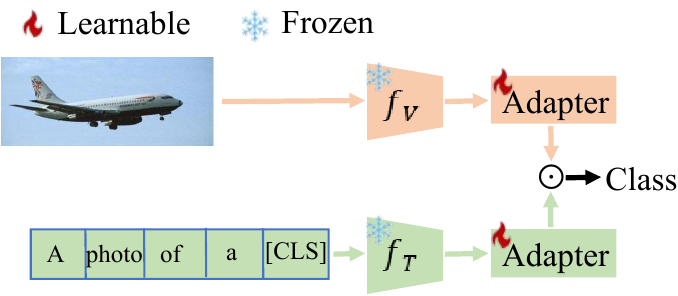}\\
    \small{(a) Traditional processing method} \\
    \includegraphics[width=0.74\linewidth]{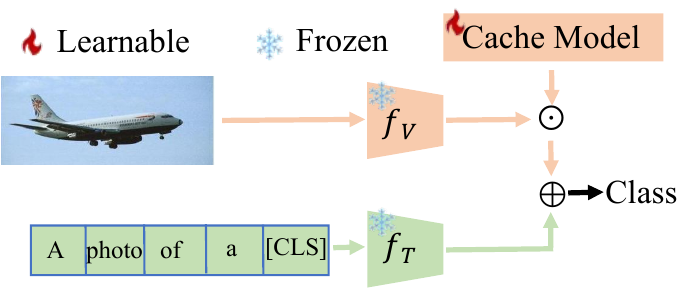}\\
    \small{(b) Training-free adaptation method}\\
    \includegraphics[width=0.74\linewidth]{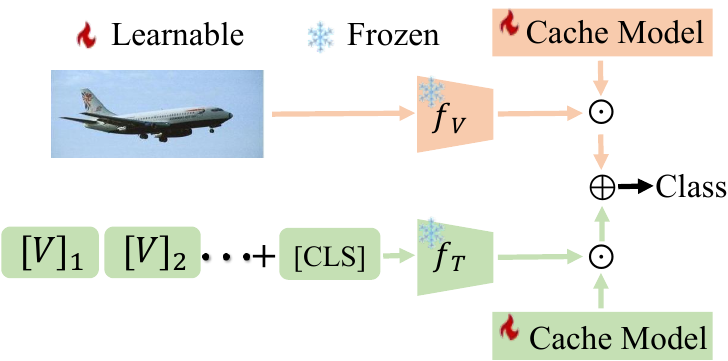}\\
    \small{(c) Our method \textbf{XMAdapter}}\\
    \vspace{2mm}
\caption{XMAdapter and Competing Methods: CLIP-Adapter~\cite{gao2023clip}, Tip-Adapter~\cite{zhang2022tip}. XMAdapter learns.}
\label{fig:introduc}
	\vspace{-3mm}
\end{figure}

The adapter technique, based on parameter fine-tuning, has shown significant effectiveness in VLM. In Fig.~\ref{fig:introduc}(a), CLIP-Adapter~\cite{gao2023clip} achieves impressive results by freezing the parameters of the visual encoder ($f_V$) and the textual encoder ($f_T$). Adapters are then added separately to $f_V$ and $f_T$ to extract information from the image and text domains. The extracted features are connected to the original features through residual connections, leading to excellent performance in image classification and recognition tasks.
In Fig.~\ref{fig:introduc}(b), Tip-Adapter~\cite{zhang2022tip}, during the training phase, stores the labels and corresponding features of images as a cache model. During the inference phase, the input image features undergo cosine similarity calculation with the stored image features in the cache model. This similarity computation, combined with the original CLIP~\cite{radford2021learning} features, aims to enhance the model's performance. GraphAdapter~\cite{li2023graphadapter} proposes modeling images and text as separate sub-graphs to enhance the performance of the model. 

These adapters are designed for either the image or the text, with the two parts working independently and without merging or interacting information. How to fully leverage the fused information between images and text has become a focal point of attention in the research community.
To address this issue, we propose a XMAdapter approach that integrates textual and image information, as illustrated in Fig.~\ref{fig:introduc}(c). The model establishes key-value pairs for both the image and text domains, embedding textual knowledge into the image domain, thereby creating a cross-modal cache model.

To further enhance the model's performance, we propose a method that independently adjusts the ratio of images and text. This is achieved by setting different ratios to adjust the fusion degree of image cache and text cache, thereby decoupling the measurement methods of similarity between different modalities. This addresses the difficulty of classifying hard samples when using only images or text. We explore hard samples based on the differences in cross-modal affinities and dynamically adjust the learning intensity for these samples, thereby further enhancing the model's performance.

During the inference stage, the model first calculates the similarity between the features of the test data and the key-value pairs in the cache model. Subsequently, it combines the model's prediction with the original CLIP's prediction through residual connections. 
Our key contributions can be summarized as follows:
\begin{itemize}
    \item We propose a novel cross-modal cache model that independently constructs cache models for images and text. This model effectively integrates features from both modalities, proving crucial for efficient transfer learning in VLM models.
    \item The model leverages retrieval through bimodal information from visual and language modalities to gather clues for inference. By dynamically adjusting the affinity ratio, it achieves cross-modal fusion and exploits the differences in modality affinity to mine hard samples. This is done through adaptive adjustments to the learning intensity of the samples, enhancing the model's performance.
    \item XMAdapter demonstrates strong representation learning capabilities, achieving excellent results on 11 benchmark datasets, including tasks such as image recognition and image classification. Furthermore, it showcases robust generalization performance on 4 benchmark datasets.
\end{itemize} 

\begin{figure*}[t]
   \center{\includegraphics[width=15cm]{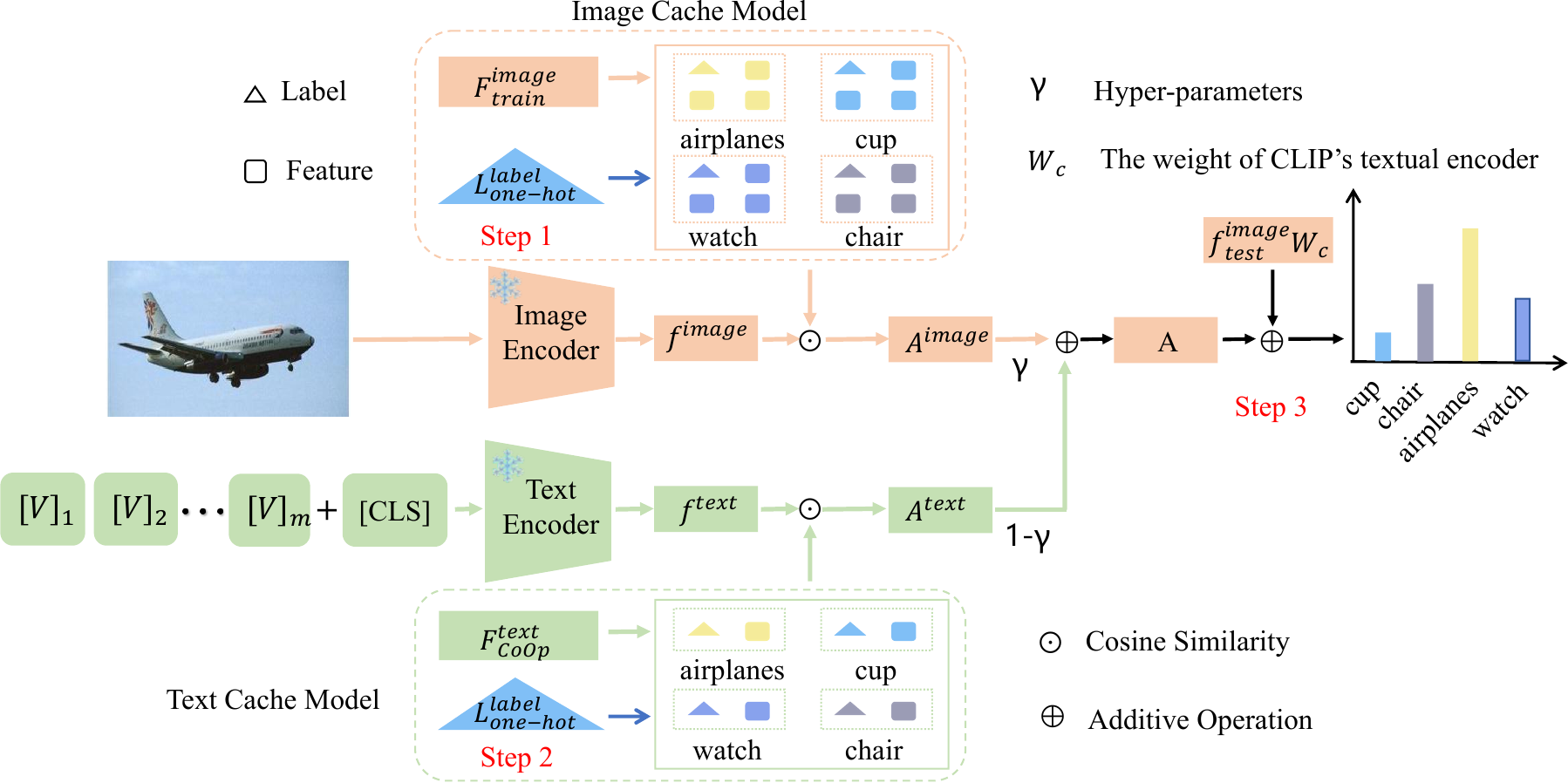}} 
   \caption{Illustration of the proposed XMAdapter. The (\textcolor[RGB]{248,203,173}{reddish orange}) line depicts the flow of image features, while the (\textcolor[RGB]{197,224,180}{pea green}) line represents the flow of text features. The model initially establishes a cache model with key-value pairs. Subsequently, it enhances the robustness of the model through adaptive adjustment of the fusion ratio between images and text, along with a strategy for learning hard samples. Finally, the model incorporates the knowledge from the original VLM to improve the accuracy of predictions.}
   \label{fig:framework}
   \vspace{-1.2em}
\end{figure*}

\section{Methodology}
\subsection{Image Cache Model Construction}
To thoroughly leverage the knowledge within the training data, we have established a key-value cache model as a feature adapter. The specific procedure is as follows: for each training image, we utilize the pre-trained CLIP visual encoder to extract a C-dimensional L2 normalized feature. Subsequently, we convert the actual label into an N-dimensional one-hot vector, denoted as $L_N$. For the K-shot N-class training samples $I_K$, where each class consists of K labeled images, resulting in a total of NK training samples, we use $F_{\text{train}}^{\text{image}}$ and $L_{\text{one-hot}}^{\text{label}}$ to represent visual features and label vectors, respectively, serving as the key and value for the cache model. This key-value pair memorizes the newly extracted knowledge from a small training dataset. Finally, the affinities of the image side, denoted as $A^{\text{image}}$, can be described as follows:
\begin{equation}
\small
\begin{aligned}
  &F_{\text{train}}^{\text{image}} = \text{ImageEncoder}(I_K),\\
  &L_{\text{one-hot}}^{\text{label}} = \text{OneHot}(L_N),\\
  &A^{\text{image}} = cos(f_{\text{test}}^{\text{image}}, F_{\text{train}}^{\text{image}}),\\
\end{aligned}
\end{equation}

where ``ImageEncoder'' is the visual encoder of CLIP, ``OneHot'' is the operation that transforms $L_N$ into a one-hot encoding, and $f_{\text{test}}^{\text{image}}$ represents the features obtained from the testing images after passing through the ``ImageEncoder''. ``cos'' denotes the calculation of the cosine similarity between the two features.

\subsection{Cross-modal Cache Model Construction}
To better utilize information across different modalities, we have devised a cross-modal cache model with the following steps: Firstly, by linearly mapping the features of the pre-trained textual side $F_{\text{CoOp}}^{\text{text}}$ from CoOp through the $\text{MetaNet}$ network to a low-dimensional space $D$, obtaining the feature vector of meta2text-feature. Subsequently, by using the values of the cache label to query the feature vector, we generate a matrix with dimensions of the total number of samples $N$ and $D$. Then, $f_\text{test}^{\text{image}}$ is mapped through the $\text{Img2TxtNet}$ network to a low-dimensional space $D$ to generate the feature vector of image2text-feature. Finally, the affinities of the textual side, denoted as $A^{\text{text}}$, can be described as follows:
\begin{equation}
\small
\label{eq:text_cache}
\begin{aligned}
    A^{\text{text}} = cos(\text{MetaNet}(F_{\text{CoOp}}^{\text{text}}),\text{Img2TxtNet}(f_\text{test}^\text{image})),
   \end{aligned}
\end{equation}
where the $\text{MetaNet}$ and $\text{Img2TxtNet}$ represents a linear neural network (MLP).

\begin{figure*}[htbp]
\centering

\subfigure
{
    \begin{minipage}[b]{0.21\linewidth}
        \centering
        \includegraphics[scale=0.21]{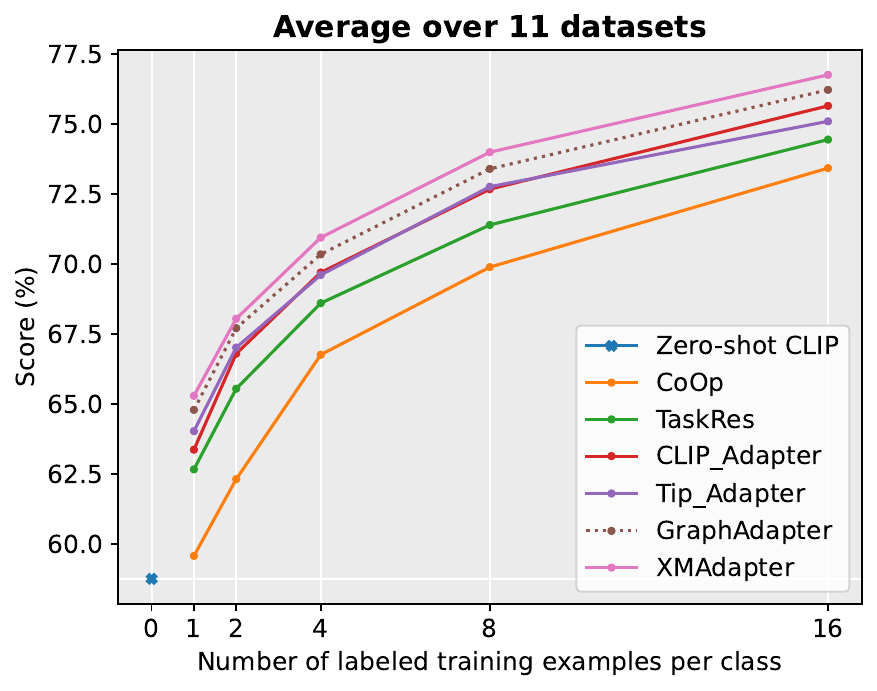}
        \label{fig:average}
    \end{minipage}
}
\subfigure
{
 	\begin{minipage}[b]{0.21\linewidth}
        \centering
        \includegraphics[scale=0.21]{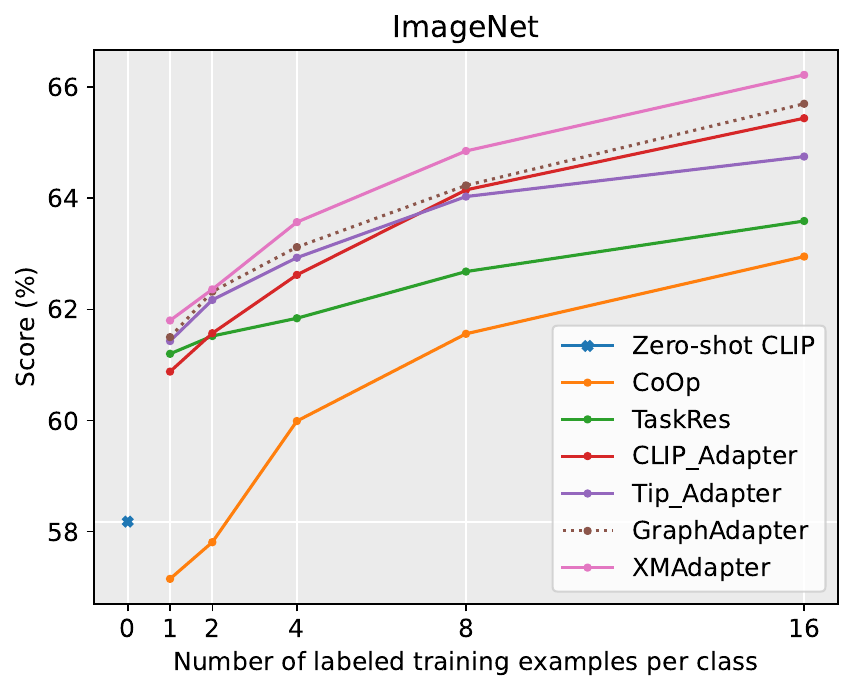}
        \label{fig:ImageNet}
    \end{minipage}
}
\subfigure
{
 	\begin{minipage}[b]{0.21\linewidth}
        \centering
        \includegraphics[scale=0.21]{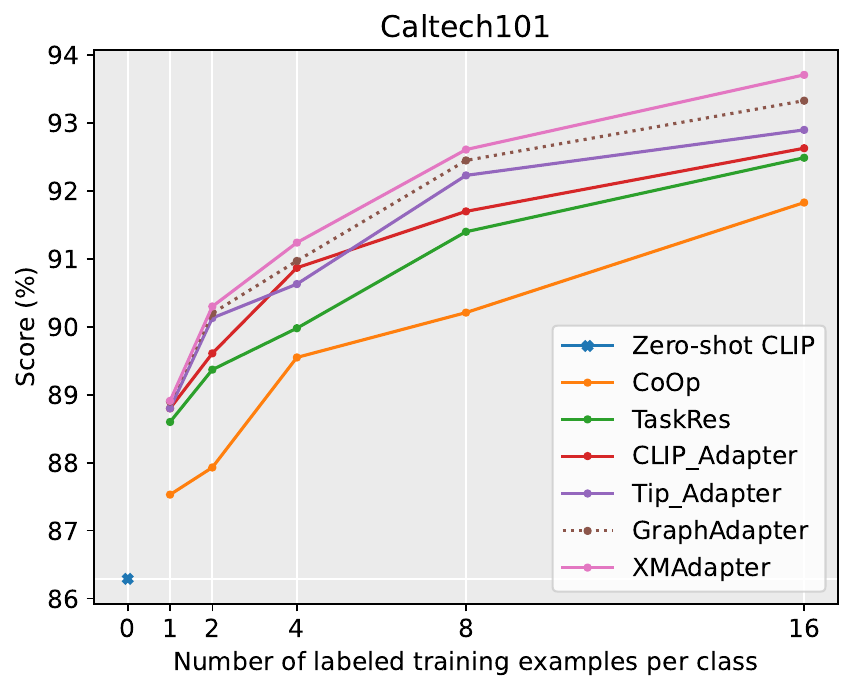}
        \label{fig:Caltech101}
    \end{minipage}
}
\subfigure
{
    \begin{minipage}[b]{0.21\linewidth}
        \centering
        \includegraphics[scale=0.21]{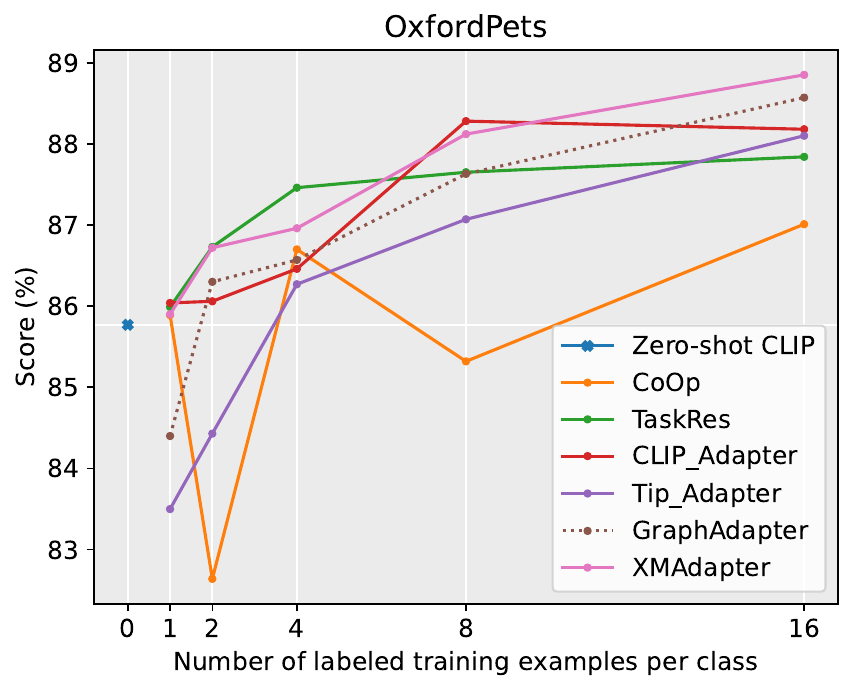}
        \label{fig:OxfordPets}
    \end{minipage}
}
\subfigure
{
 	\begin{minipage}[b]{0.21\linewidth}
        \centering
        \includegraphics[scale=0.21]{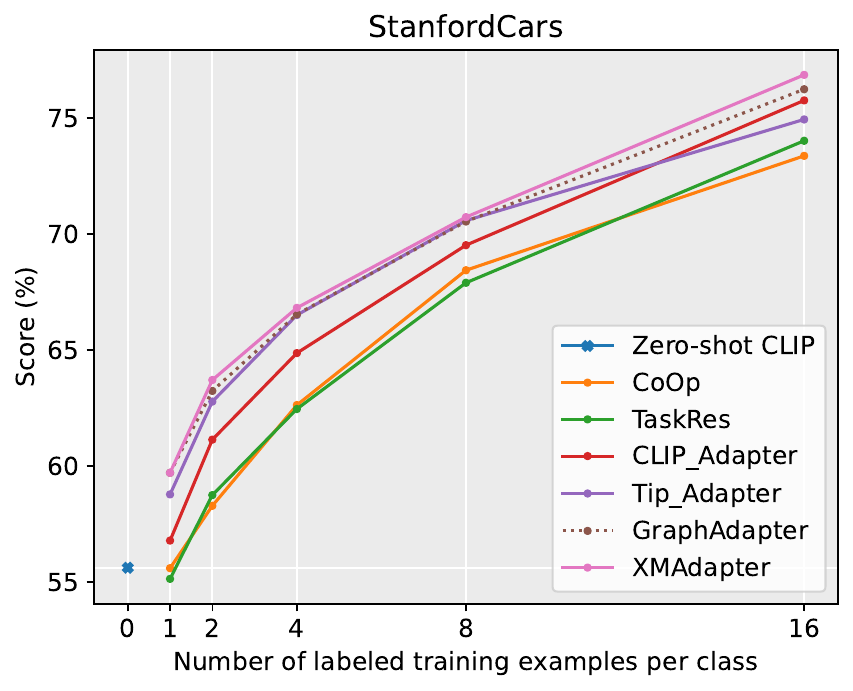}
        \label{fig:StanfordCars}
    \end{minipage}
}
\subfigure
{
 	\begin{minipage}[b]{0.21\linewidth}
        \centering
        \includegraphics[scale=0.21]{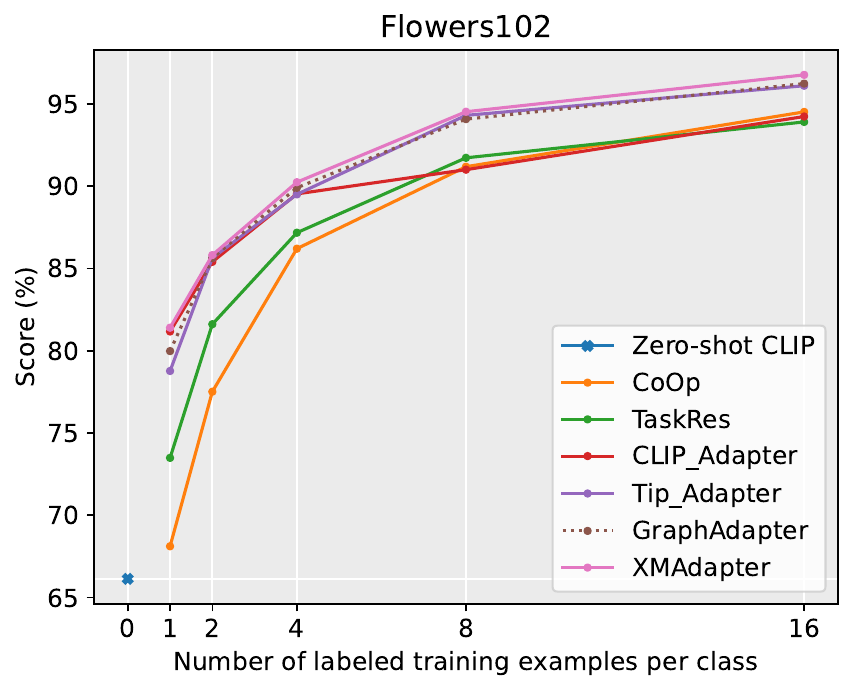}
        \label{fig:Flowers102}
    \end{minipage}
}
\subfigure
{
    \begin{minipage}[b]{0.21\linewidth}
        \centering
        \includegraphics[scale=0.21]{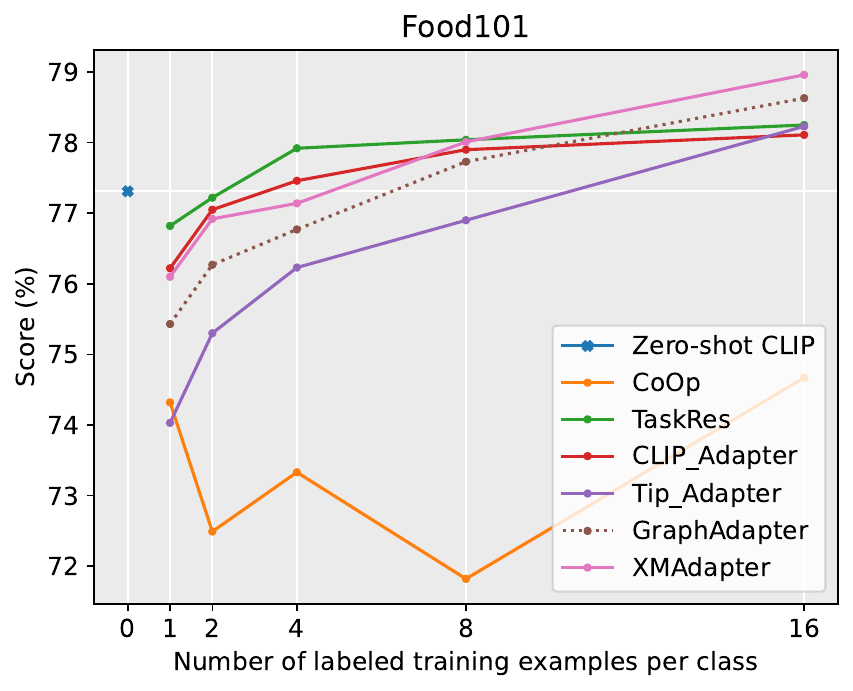}
        \label{fig:Food101}
    \end{minipage}
}
\subfigure
{
 	\begin{minipage}[b]{0.21\linewidth}
        \centering
        \includegraphics[scale=0.21]{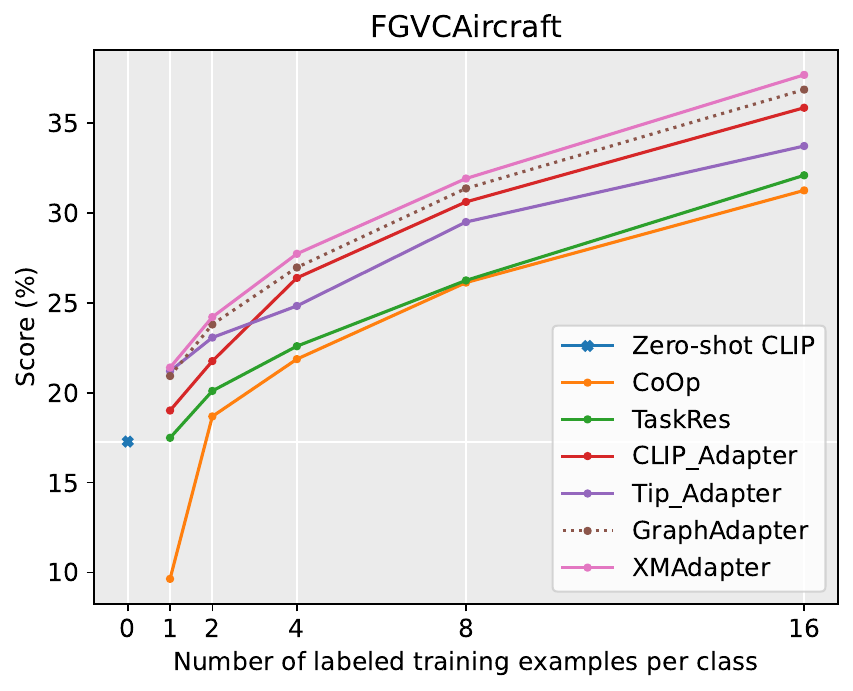}
        \label{fig:FGVCAircraft}
    \end{minipage}
}
\subfigure
{
 	\begin{minipage}[b]{0.21\linewidth}
        \centering
        \includegraphics[scale=0.21]{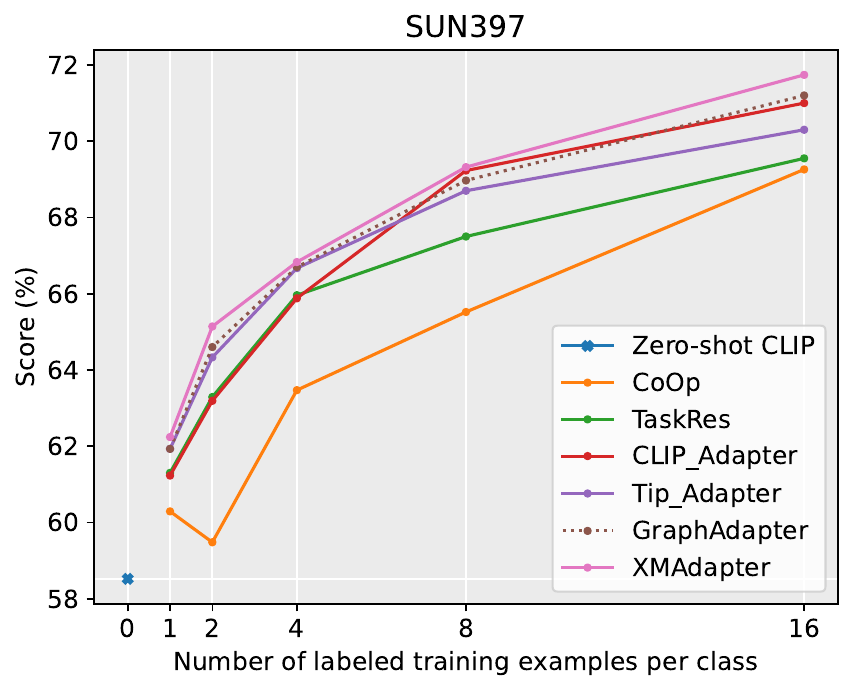}
        \label{fig:SUN397}
    \end{minipage}
}
\subfigure
{
    \begin{minipage}[b]{0.21\linewidth}
        \centering
        \includegraphics[scale=0.21]{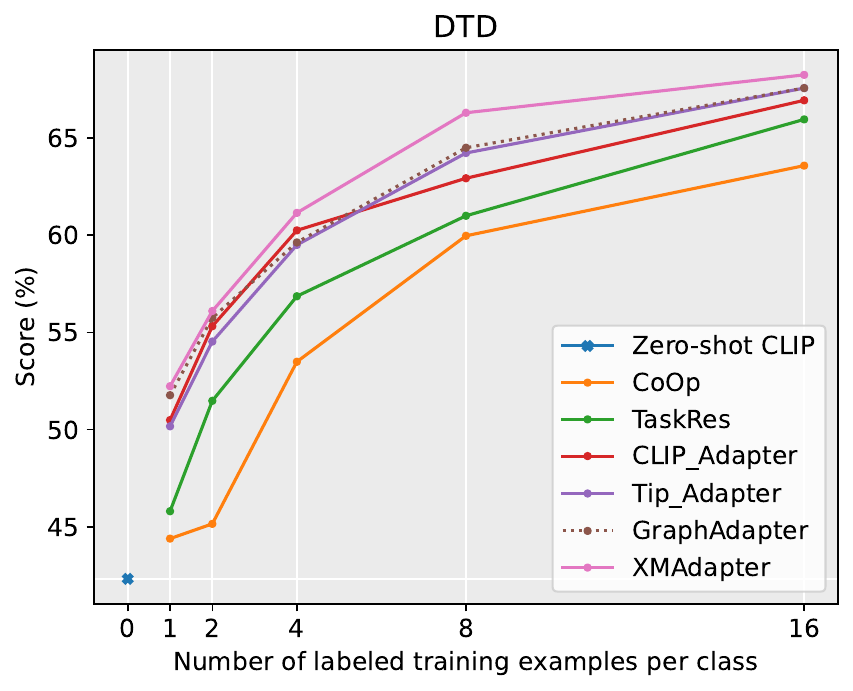}
        \label{fig:DTD}
    \end{minipage}
}
\subfigure
{
 	\begin{minipage}[b]{0.21\linewidth}
        \centering
        \includegraphics[scale=0.21]{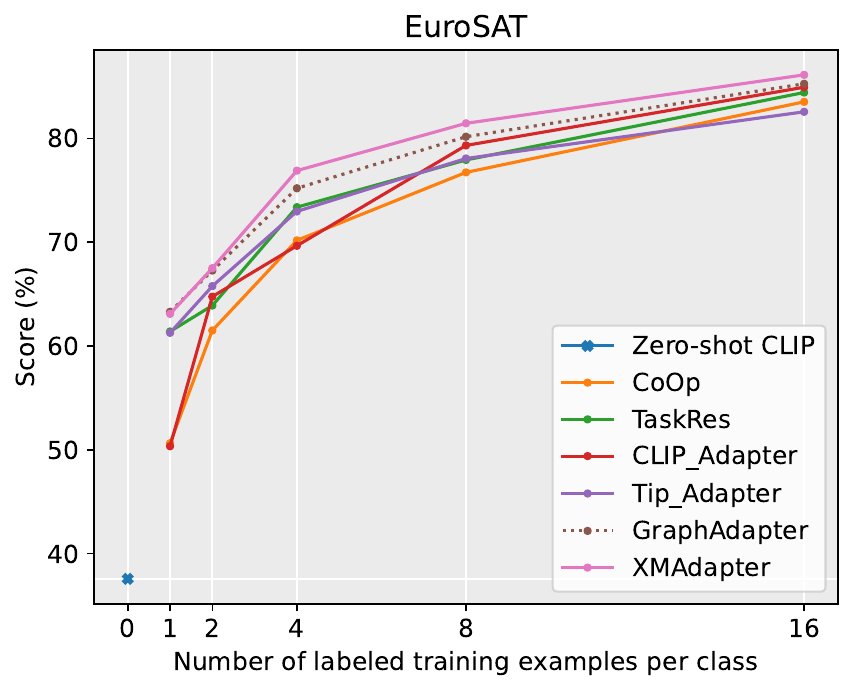}
        \label{fig:EuroSAT}
    \end{minipage}
}
\subfigure
{
 	\begin{minipage}[b]{0.21\linewidth}
        \centering
        \includegraphics[scale=0.21]{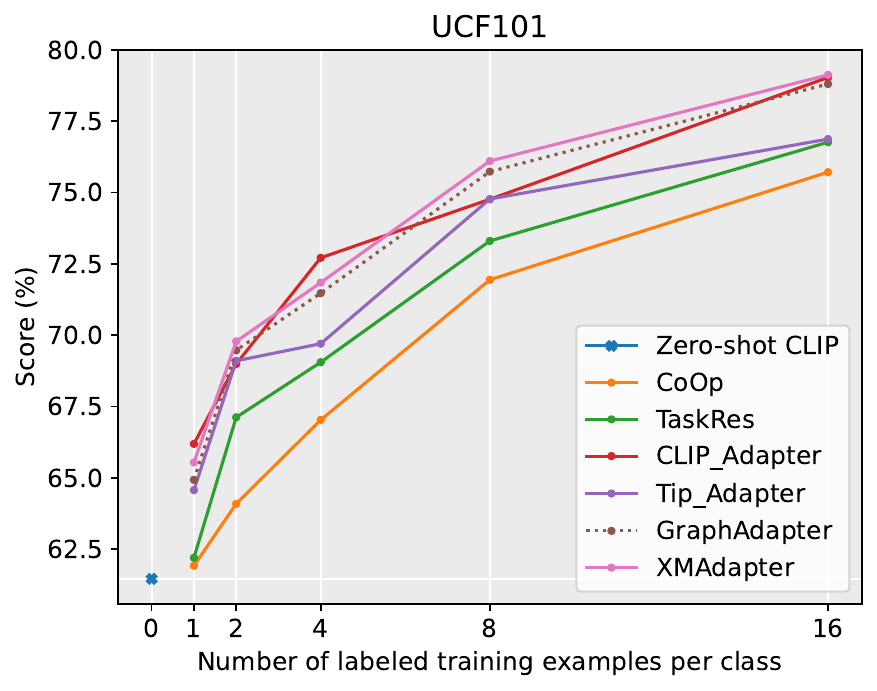}
        \label{fig:UCF101}
    \end{minipage}
}
\caption{The performance comparison of our XMAdapter with the SOTA method on Cross Label Generalization, including 1-/2-/4-/8-/16-shots on 11 benchmark datasets.}
\label{fig:crosslable}
\end{figure*}

\subsection{Adaptive Scaling}
To further enhance the model's performance, we separately calculate the similarities on the image and text sides, To better understand the contribution of different modalities to the model's classification results, we propose a dynamic adjustment method for the proportions of $A^{\text{image}}$ and $A^{\text{text}}$, decoupling the measurement methods of similarity between different modalities. The adjustment process is as follows:
\begin{equation}
\label{eq:gamma}
\small
\begin{aligned}
   A = \gamma A^{\text{image}} + (1-\gamma) A^{\text{text}},
\end{aligned}
\end{equation}
where $\gamma$ represents the adaptive adjustment coefficient.

\subsection{Building Logits}
The model acquires knowledge from two sources. One part comes from the cache model constructed with a small number of labeled samples, which is obtained by multiplying the affinities matrix $A$ by $L_{\text{one-hot}}^{\text{label}}$. The other part comes from the prior knowledge of the original CLIP classifier $W_c$. The contributions of these two terms are balanced by the weight $\alpha$. The entire process of constructing logits can be described as follows:
\begin{equation}
\small
\label{eq:beta}
\begin{aligned}
   & \text{logits}_\text{cache} = exp(- \beta (1-AL_{\text{one-hot}}^{\text{label}} )), \\
   & \text{logits} = \alpha \text{logits}_\text{cache} + f_\text{test}^{\text{image}} W_c, 
   \end{aligned}
\end{equation}
where the $\alpha$ controls the fusion ratio and the $\beta$ controls the sharpness of the affinities matrix.

\subsection{Hard Example Mining}
The model performs well under the cache model. To further improve its performance, we adopt the OHEM (Online Hard Example Mining) approach. Specifically, for hard example samples, we set different weights to enhance the model's accuracy. The process is as follows: We obtain the affinities matrix $A^{\text{image}}$ for images and the affinities matrix $A^{\text{text}}$ for text. By leveraging the differences in affinity between modalities, we aim to identify hard samples. The weights for learning hard examples are adaptively adjusted and can be described as follows:
\begin{equation}
  \small
    \begin{aligned}
        A_b^{\text{weight}} =
        \frac{1}{N}\sum_{n=1}^N \text{sigmod}(\|A_{bn}^{\text{image}} - A_{bn}^{\text{text}} \|),
   \end{aligned}
\end{equation}
where $\|.\|$ denotes the absolute value of two numbers, $A_{bn}^{\text{image}}$ is a sample in the $A^{\text{image}}$ affinities matrix, $N$ represents the number of samples in the cache model, and $\text{sigmoid}$ is the threshold function with values ranging between [0,1].

During the training phase, the model first calculates the cross-entropy loss $\mathcal{L}_{b}^{ce}$ between logits and the labels of the training samples. Subsequently, it adjusts the loss by incorporating the mean of $A_b^{\text{weight}}$ to form the final loss function. The process can be described as follows:
\begin{equation}
\small
\label{eq:A_ce_loss}
    \begin{aligned}
        & \mathcal{L}_{b}^{ce} = -\frac{1}{K}\sum_{k=1}^K y_k log(\text{logits}_k),\\
        & \mathcal{L} = 
        \frac{1}{B}\sum_{b=1}^B \mathcal{L}_{b}^{ce} * A_b^{\text{weight}},
    \end{aligned}
\end{equation}
where $K$ is the number of classes, $y$ represents the label of the sample, $B$ is the total number of samples in a batch, and $b$ represents a sample in the batch.

\section{Experiment}
\begin{table*}
\centering
\small
\caption{The performance comparison regarding generalization capability on four CLIP visual backbones. The ETL methods are optimized with the ImageNet dataset on a 16-shot setting and tested on cross-domain datasets, including ImageNet-V2, ImageNet-Sketch, ImageNet-A, and ImageNet-R.}
\resizebox{0.6\textwidth}{!}{\begin{tabular}{lcccccccc}
\hline
\multirow{2}{*}{Method}            &  \multirow{2}{*}{Backbone} & Source &  & \multicolumn{5}{c}{Target}          \\  \cline{3-3} \cline{5-9}
 &  & ImageNet & & -V2 & -Sketch & -A & -R & Average 
\\ \hline
Zero-shot CLIP~\cite{radford2021learning}    & \multirow{6}{*}{ResNet-50}  &  58.18  &      &   51.34          &   33.32              &   21.65         &   56.00         &  40.58       \\
Linear Probe CLIP~\cite{radford2021learning} &                             &   55.87 &     &      45.97       &  19.07                &   12.74        &  28.16          &    26.49     \\
CoOp~\cite{zhou2022learning}              &                           &      62.95  &    &    55.11        & 32.74                 &     22.12      &   54.96         & 41.23           \\
TaskRes~\cite{yu2023task}           &                             &     64.75 &   &    56.47           & 35.83                 &     22.80         &      60.70      &    43.95      \\ 
GraphAdapter~\cite{li2023graphadapter}  &                          &   65.70    &     &   56.40         &   34.50          & 21.88  & 58.94  &  42.93 \\
\rowcolor[gray]{0.9} XMAdapter  &                          &   \textbf{66.22}    &     &\textbf{56.51}         &   \textbf{36.72}              &        \textbf{23.46} 
&    \textbf{61.53}        &  \textbf{44.56} 
\\ \hline

Zero-shot CLIP~\cite{radford2021learning}    & \multirow{6}{*}{ResNet-101} &              61.62  &    &  54.81            &     38.71     &          28.05     &64.38       & 46.49        \\
Linear Probe CLIP~\cite{radford2021learning} &                             &    59.75  &    &    50.05        &   26.80                 &     19.44        &    47.19          &  35.87       \\
CoOp~\cite{zhou2022learning}               &                             &   66.60 &    &   58.66           &        39.08          &  28.89          &       63.00      &  47.41        \\
TaskRes~\cite{yu2023task}          &                             & 67.70    &     &     59.50         &          \textbf{41.70}         &     29.87        &     68.07        &   49.79      \\
GraphAdapter~\cite{li2023graphadapter}   &                             & 68.23   &      & 59.60         & 40.83               & 28.77    & 67.13          & 49.08     \\ 
\rowcolor[gray]{0.9} XMAdapter   &                             & \textbf{68.96}   &      & \textbf{59.64}         &  41.50               & \textbf{30.57}    &  \textbf{68.82}          & \textbf{50.13}     \\ 
\hline

Zero-shot CLIP~\cite{radford2021learning}    & \multirow{6}{*}{ViT-B/32}   &   62.05   &     &   54.79        & 40.82                    & 29.57      &   65.99           &   47.79      \\
Linear Probe CLIP~\cite{radford2021learning} &                             &   59.58   &     &     49.73         &    28.06               &    19.67       &   47.20          &  36.17         \\
CoOp~\cite{zhou2022learning}              &                             &   66.85 & &   58.08            & 40.44                     &    30.62          &    64.45          &    48.40     \\
TaskRes~\cite{yu2023task}           &                             &      68.20 &&     \textbf{59.20}         & 42.50                   &    31.43          &    69.33          &     50.62     \\
GraphAdapter~\cite{li2023graphadapter}             &                             &     68.80  &  &    59.00        &             41.70    &   29.57        &  68.67          &  49.74 \\
\rowcolor[gray]{0.9} XMAdapter            &                             &     \textbf{69.56}  &  &    59.12        &             \textbf{42.91}    &   \textbf{31.95}        &  \textbf{69.57}          &  \textbf{50.89} 
\\ \hline

Zero-shot CLIP~\cite{radford2021learning}    & \multirow{6}{*}{ViT-B/16}   & 66.73  &       &  60.83         &   46.15                  & 47.77         &     73.96            & 57.18         \\
Linear Probe CLIP~\cite{radford2021learning} &                             &  65.85   &      &    56.26       &   34.77                  &    35.68        &  58.43        &    46.29        \\
CoOp~\cite{zhou2022learning}             &                             &  71.92 &  &   64.18           & 46.71                      &     48.41       &    74.32           &  58.41       \\
TaskRes~\cite{yu2023task}           &                             &      73.07  & &  65.30      &  49.13                        &   50.37    &            77.70 & 60.63           \\
GraphAdapter~\cite{li2023graphadapter}              &                             &   73.68  &    &   \textbf{65.57}        &             48.57    &   49.23       & 77.20          &  60.14 \\
\rowcolor[gray]{0.9} XMAdapter             &                             &   \textbf{74.43}  &    & 65.54        &             \textbf{49.58}    &   \textbf{50.69}       & \textbf{77.95}          & \textbf{60.94}  
\\ \hline
\end{tabular}}
\label{tab:generalization}
\vspace{-2mm}
\end{table*}

\subsection{Cross Label Generalization}
We compared the performance of XMAdapter with Zero-shot CLIP~\cite{radford2021learning}, CoOp~\cite{zhou2022learning}, TaskRes~\cite{yu2023task}, CLIP-Adapter~\cite{gao2023clip}, Tip-Adapter~\cite{zhang2022tip}, and GraphAdapter~\cite{li2023graphadapter} on 11 datasets, as shown in Fig.~\ref{fig:crosslable}. The model performed well across all 11 datasets in 1-/2-/4-/8-/16-shots settings. Particularly, at 16 shots, XMAdapter achieved an average accuracy of 76.87\% across the 11 datasets, surpassing GraphAdapter's 76.22\% by 0.65\%. On the challenging fine-grained classification dataset, FGVCAircraft, XMAdapter outperformed the six compared methods in the 2-/4-/8-/16-shots settings. This demonstrated that the cross-modal adapter design of XMAdapter, integrating both image and text information, is better suited for downstream tasks. 
A detailed introduction to the comparative methods can be found in the Appendix~\ref{app:relate}.

\subsection{Domain Generalization}
We further tested the generalization capability of the XMAdapter, and the experimental results are shown in Table~\ref{tab:generalization}. The model is trained with 16-shot training samples using ImageNet~\cite{deng2009imagenet} as the training dataset. The testing datasets include ImageNet-V2, ImageNet-Sketch, ImageNet-A, and ImageNet-R. These testing datasets share the same categories as ImageNet~\cite{deng2009imagenet} but differ in background, texture, semantics, and other aspects. We used ResNet-50~\cite{he2016deep}, ResNet-101~\cite{he2016deep}, ViT-B/32~\cite{DBLP:conf/iclr/DosovitskiyB0WZ21}, and ViT-B/16~\cite{DBLP:conf/iclr/DosovitskiyB0WZ21} as backbones. XMAdapter achieved an average improvement of +0.61\%, +0.34\%, +0.27\%, and +0.31\% on four datasets, respectively. The experimental results demonstrated that the XMAdapter exhibits strong generalization capabilities with its cross-modal cache model. 
For details regarding the datasets and implementation details in the Appendix~\ref{app:exp}.

\begin{table*}[t]
    \label{tab1}
	\centering
	\caption{Comparison of classification accuracy (\%), time efficiency, parameters, and GFlops for different methods on 16-shot ImageNet~\cite{deng2009imagenet}, where our proposed XMAdapter achieve superior accuracy-efficiency trade-off.}
	\adjustbox{width=0.8\linewidth}{
	\begin{tabular}{lccccccc}
	\toprule
		Models                           &Tunable Parameters(M) &GFlops   &Training Time(one epoch)(s) &Inference Time &GPU Memory &Performance  \\ \midrule
		CoOp~\cite{zhou2022learning}     & \textbf{0.008}       &1943.12  & 40.91        &119.64ms       &18.907  &62.95\\
		CLIP-Adapter~\cite{gao2023clip}  & 0.524                &1959.44  & 45.71        &275.22ms       &9.257   &63.59\\ 
	    Tip-Adapter~\cite{zhang2022tip}  & 16.384               &5.43    &\textbf{12.36} &51.03ms        &\textbf{4.313}   &65.44\\
        TaskRes~\cite{yu2023task}        & 1.024                &5.42     & 13.64   &\textbf{4.89ms}     &6.227   &64.75\\
  GraphAdapter~\cite{li2023graphadapter} & 4.145                &5.42     & 23.29        &4.91ms         &10.75   &65.70\\ \midrule
	                     XMAdapter  & 18.561      &\textbf{5.39}     & 13.41        &73.24ms        &5.148   &\textbf{66.22} \\ 
	\bottomrule
	\end{tabular}}
\label{tab:modelcomple}
\end{table*}

\subsection{Model Complexity}
We compared our experimental results with existing efficient transfer learning methods from six perspectives: tunable parameters, GFlops, training time, inference time, GPU memory, and performance. All experiments were tested with batch size 32 on a single NVIDIA GeForce RTX 3090 GPU. The experiments were conducted under the setting of 16-shot on the ImageNet~\cite{deng2009imagenet} dataset, and the results are presented in Table~\ref{tab:modelcomple}. We observed that the tunable parameters of the XMAdapter were slightly higher than those of the Tip-Adapter~\cite{zhang2022tip}, mainly due to the more intricate process involved in establishing the cross-modal cache model compared to the conventional cache model. The GFlops value was comparable to Tip-Adapter~\cite{zhang2022tip}, TaskRes~\cite{yu2023task}, and GraphAdapter~\cite{li2023graphadapter}. The training time was lower than that of CoOp~\cite{zhou2022learning}, TaskRes~\cite{yu2023task}, and GraphAdapter~\cite{li2023graphadapter}. In terms of inference time and GPU memory, the XMAdapter ranks in the middle among the compared methods. Furthermore, the model's performance is 3.27\% higher than CoOp~\cite{zhou2022learning} and 0.52\% higher than GraphAdapter~\cite{li2023graphadapter}. Therefore, the XMAdapter meets the requirements of parameter-efficient transfer learning in terms of resource consumption, and operational efficiency, and achieves promising experimental results.

\subsection{Ablation Studies}
\textbf{Different Hyper-parameters $\gamma$.}
In Equation~\ref{eq:gamma}, $\gamma$ is the parameter adjusting the fusion ratio of $A^{\text{image}}$ and $A^{\text{text}}$. 
To determine an appropriate value for $\gamma$, we conducted tests on the 16-shot ImageNet dataset, as shown in Table~\ref{lib:gamma-scaling}. When $\gamma$ is set to 0, only $A^{\text{text}}$ influences the model, making it similar to CoOp~\cite{zhou2022learning}, achieving an accuracy of 62.95. When $\gamma$ is set to 1, the model is similar to the configuration of Tip-Adapter~\cite{zhang2022tip}, reaching an accuracy of 65.44. The best result of 66.22 is achieved when $\gamma$ is set to 0.7, further validating the rationality of XMAdapter in fusing multiple features.

\begin{table}
\footnotesize
\centering
\caption{Comparing the effect of the $\gamma$ on 16-shot ImageNet.}
\begin{adjustbox}{width=0.85\linewidth}
    \begin{tabular}{cccccccc}
    \toprule
    \small
        $\gamma$ & 0 & 0.1 & 0.3 & 0.5 & 0.7 & 0.9 & 1.0 \\ \midrule
        Acc   & 62.95 & 64.44  & 65.13 & 65.95 & \textbf{66.22} & 65.87 & 65.44 \\ \bottomrule
    \end{tabular}
   \label{lib:gamma-scaling}
   \vspace{-0.8em}
\end{adjustbox}
\end{table}

\textbf{Different coefficient $\alpha$ and $\beta$.}
$\alpha$ represents the proportion of the prediction in the cross-modal cache model combined with pre-trained CLIP, as illustrated in Eq.~\ref{eq:beta}. 
To assess the impact of different coefficients $\alpha$ on model performance, we conducted tests on the 16-shot ImageNet dataset. We assumed a range of values for $\alpha$ from 0 to 4.0, and the experimental results are presented in Table~\ref{tab:alphabeta}. When $\alpha$ is set to 0, XMAdapter essentially utilizes only the knowledge from pre-trained CLIP, disregarding the content from the cache model. The model achieves a performance of 58.18\%. When $\alpha$ is set to 1.2, the model achieves its best performance of 66.22\%. However, when $\alpha$ is increased to 4.0, the model's performance decreases. Experimental results indicate that the knowledge from the cache model and pre-trained CLIP hold equal importance in the model. As indicated in Eq.~\ref{eq:beta}, When $\beta$ increases, it signifies that similar samples in the test images have a greater impact on the model's predictions. As shown in Table~\ref{tab:alphabeta} hold, the model achieves its best performance when $\beta$ is set to 3.5. 
For details regarding the performance of the model on different backbones, please refer to Different BackBones in the Appendix~\ref{sub:diff_backbone}.

\begin{table}[t]
\centering
\caption{On the 16-shot ImageNet, we examined the impact of the two different coefficients, namely, the residual ratio $\alpha$ and the sharpness ratio $\beta$, on XMAdapter.
}
\begin{adjustbox}{width=0.85\linewidth}
	\begin{tabular}{c|ccccccc}
	\toprule
 
		\multicolumn{8}{c}{Ablation Studies on XMAdapter} \\ 
		\midrule
  
		\multirow{2}{*}{\shortstack{Residual Ratio\ \ \\$\alpha$}}
	   &0.0 &0.5 &1.0 &1.2 &\ \ 2.0\  &\ 3.0\ \ &\ 4.0 \\ 
        \cmidrule(lr){2-8}
		 &\ 58.18  & 64.57 & 65.73 &\textbf{66.22} &\ \ 65.42 & \ 63.83 &\ 61.37\\ 
        \midrule
		
	    \multirow{2}{*}{\shortstack{Sharpness Ratio\ \  \\$\beta$}} 
           &0.5 &1.5 &3.5 &5.5 &7.5 &9.5 &11.5 \\
	     \cmidrule(lr){2-8}
	     &\ 64.37  &64.85 &\textbf{66.22} &65.03 &\ \ 64.64 &\ 64.26 &\ 65.97\\
	\bottomrule
	\end{tabular}
\end{adjustbox}
\label{tab:alphabeta}
\end{table}

\section{Conclusion}
In this paper, we first analyze the drawbacks of current adapter-style methods in parameter-efficient transfer learning applications. Subsequently, we introduce XMAdapter as a solution, which creates a cache model by integrating cross-modal information to acquire knowledge. It dynamically adjusts the weights of hard examples based on the differences in affinity between modalities to enhance model performance. We validated the effectiveness and generalization of the model on 15 benchmark datasets. 
Adapter-style methods are based on fine-tuning VLM with a small amount of data to match downstream tasks. This tuning approach heavily relies on the performance of the pre-trained VLM.

\bibliographystyle{IEEEbib}
\bibliography{icme2023template}

\newpage
\appendix
\section*{Appendix}
In this supplementary material, we first provide the related work in APPENDIX~\ref{app:relate} and the experiment in APPENDIX~\ref{app:exp}.

\section{Related Work}
\label{app:relate}
\subsection{Vision-Language Model}
Influenced by the tremendous success of pre-trained models in computer vision and natural language processing, the community has begun applying pre-training techniques to vision and language models. Classic Vision-Language Models typically consist of a vision encoder, a language encoder, a fusion encoder, and a loss function. In the early stages, models like BAN~\cite{kim2018bilinear}, MCAN~\cite{yu2019deep}, and Intra-Inter~\cite{gao2019dynamic} dominated the scene. Influenced by the BERT~\cite{DBLP:conf/naacl/DevlinCLT19} philosophy, models like LXMERT~\cite{DBLP:conf/emnlp/TanB19}, ViLBERT~\cite{lu2019vilbert}, and UNITER~\cite{DBLP:journals/corr/abs-1909-11740} have achieved remarkable success in the Vision-Language Model domain. Recently, models such as CLIP~\cite{radford2021learning}, DeCLIP~\cite{DBLP:conf/iclr/LiLZCOSYY22}, BLIP~\cite{li2022blip}, and ALIGN~\cite{jia2021scaling} have demonstrated that contrasted learning based on visual and language inputs can generate transferable features. They achieve good performance on downstream tasks without the need for fine-tuning. CoOp~\cite{zhou2022learning} transformed the fixed prompts in CLIP into learnable prompts, further enhancing the model's performance. CoCoOp~\cite{zhou2022conditional}, by incorporating the characteristics of input images, improved the model's generalization by introducing inductive biases through image augmentation. Our XMAdapter explores the model's potential by adaptively adjusting the fusion ratio between images and text, achieving good performance in downstream tasks such as image classification and image recognition.
\subsection{Parameter-Efficient Transfer Learning}
Prompt learning is widely used in model tuning, especially for large language models. P-tuning~\cite{liu2022p} proposed to make the prompt into a token and use BiLSTM for learning. P-tuning v2~\cite{liu2021p} combined the advantages of P-tuning and Prefix-tuning to make the model suitable for small models and complex natural language understanding tasks by removing repetitive parameters and enabling multi-task learning.
Inspired by prompts in the NLP field, VPT~\cite{jia2022visual} introduced a small number of trainable parameters as image prompts and achieved good results in downstream tasks. To improve the efficiency of prompt design, NOAH~\cite{zhang2022neural} used an evolutionary search technique to find the optimal design of vision prompts, adapters, and LoRA~\cite{hu2021lora} as parameter-efficient tuning modules in each layer of the Vision Transformer. DP~\cite{yang2023dynamic} proposed a method to dynamically adjust the prompt's position and length according to the instance's different tasks. UPT~\cite{zang2022unified} and MAPLE~\cite{khattak2023maple} proposed adding learnable contextual markers in the language and visual branches and mapping language prompts to visual prompts through aggregation functions.

\subsection{Cache Model}
The cache model is a collection of key-value pairs storing training data and labels. It is established during the model's training phase and inference and aggregates information from the cache model by treating the test samples as query conditions and utilizing similarity retrieval. This approach does not require updating model parameters and can improve the system's inference speed. It has been widely applied in Unbounded Cache~\cite{grave2017unbounded}, Matching Networks~\cite{vinyals2016matching}, Prototypical Networks~\cite{snell2017prototypical}, and MAML~\cite{finn2017model}. Unbounded Cache~\cite{grave2017unbounded} expanded the scale of a continuous cache through approximate nearest neighbor search and quantization algorithms. Matching Network~\cite{vinyals2016matching} adapted to new class types by establishing a small labeled support set and mapping an unlabeled example to its label. 
The XMAdapter proposed in this paper fully integrates features from both images and text, decoupling the similarity measurement methods of different modalities. This enables the mutual utilization of knowledge between the two modalities, achieving good performance in downstream tasks.

\section{Experiment}
\label{app:exp}
\subsection{Experimental Setups}
\textbf{Datasets} Following previous adapter-style studies~\cite{gao2023clip,zhang2022tip}, we validate our XMAdapter on 11 few-shot classification tasks, including ImageNet~\cite{deng2009imagenet}, Caltech101~\cite{fei2004learning}, OxfordPets~\cite{parkhi2012cats}, StandfordCars~\cite{krause20133d}, Flowers102~\cite{nilsback2008automated}, Food101~\cite{bossard2014food}, FGVCAircraft~\cite{maji2013fine}, SUN397~\cite{xiao2010sun}, DTD~\cite{cimpoi2014describing}, EuroSAT~\cite{helber2019eurosat}, and UCF101~\cite{soomro2012dataset}. Among them, OxfordPets~\cite{parkhi2012cats}, StanfordCars~\cite{krause20133d}, Flowers102~\cite{nilsback2008automated}, FGVCAircraft~\cite{maji2013fine}, and Food101~\cite{bossard2014food} belong to fine-grained classification tasks, DTD~\cite{cimpoi2014describing} is the dataset of texture classification and EuroSAT~\cite{helber2019eurosat} is for remote sensing classification. To investigate the generalization capability of our XMAdapter, we conduct experiments on ImageNetV2~\cite{recht2019imagenet}, ImageNet-Sketch~\cite{wang2019learning}, ImageNet-A~\cite{hendrycks2021natural} and ImageNet-R~\cite{hendrycks2021many}.

\textbf{Implementation Details}
This paper employs CLIP~\cite{radford2021learning} as the backbone. By default, we utilize ResNet-50~\cite{he2016deep} as the visual encoder and a 12-layer transformer as the textual encoder. To evaluate the adaptability of our method, XMAdapter also uses other CLIP~\cite{radford2021learning} visual encoders, including ResNet-101~\cite{he2016deep}, ViT-B/32~\cite{DBLP:conf/iclr/DosovitskiyB0WZ21}, and ViT-B/16~\cite{DBLP:conf/iclr/DosovitskiyB0WZ21}. We use a batch size of 32 for all datasets. We iterate for 20 epochs and optimize the model using 1, 2, 4, 8, and 16 shots. During training, we use the Adam optimizer with an initial learning rate of $1\times10^{-3}$, which decreases with cosine learning rate decay. The text in CoOp~\cite{zhou2022learning} is randomly initialized from a Gaussian distribution with a standard deviation of 0.02, and the context length is set to 16. The data augmentation strategy includes only two policies: ``random resizing crop'' and ``random flipping''.

\subsection{Different BackBones}
\label{sub:diff_backbone}
We utilized ResNet-50~\cite{he2016deep}, ResNet-101~\cite{he2016deep}, ViT-B/32~\cite{DBLP:conf/iclr/DosovitskiyB0WZ21}, and ViT-B/16~\cite{DBLP:conf/iclr/DosovitskiyB0WZ21} as backbones for conducting comparative experiments on the 16-shot ImageNet~\cite{deng2009imagenet} dataset. From Table~\ref{tab:backbone}, it can be observed that XMAdapter, across the four backbones, outperforms GraphAdapter by +0.52\%, +0.40\%, +0.76\%, and +0.74\%, respectively. This indicates that the model exhibits strong generalizability.
\begin{table}[t!]
\centering
\caption{Classification accuracy ($\%$) of different visual encoders on 16-shot ImageNet~\cite{deng2009imagenet}.}
\begin{adjustbox}{width=0.85\linewidth}
	\begin{tabular}{lccccccc}
	\toprule
		Models &\ ResNet-50\ &\ ResNet-101\ &\ ViT-B/32\ & \ ViT-B/16\ \\ \midrule
		Zero-shot CLIP~\cite{radford2021learning} &58.18 &61.62 &62.05 &66.73 \\
		CoOp~\cite{zhou2022learning} & 62.95 &66.60 & 66.85  &71.92 \\
		CLIP-Adapter~\cite{gao2023clip} & 63.59 & 65.39 & 66.19 &71.13 \\ 
		Tip-Adapter~\cite{zhang2022tip} & 65.44 & 68.56 & 68.65 & 73.69 \\
        GraphAdapter~\cite{li2023graphadapter} & 65.70 & 68.23 & 68.80 & 73.68 \\ \midrule
        XMAdapter & \textbf{66.22} & \textbf{68.96} & \textbf{69.56} & \textbf{74.43} \\
	\bottomrule
	\end{tabular}
\end{adjustbox}
\label{tab:backbone}
\end{table}

\end{document}